\documentclass[sigconf, nonacm, 10pt]{acmart}

\begin{document}
\title{2nd Place Solution to Google Landmark Recognition Competition 2021}
\fancyhead{}
\author{Shubin Dai}
\email{bestfitting.ai@gmail.com}


\begin{abstract}
As Transformer-based architectures have recently shown encouraging progresses in computer vision. In this work, we present the solution to the Google Landmark Recognition 2021 Challenge~\cite{landmark-recognition-2021} held on Kaggle, which is an improvement on our last year's solution~\cite{2ndPlaceGLR2020} by changing three designs, including
(1) Using Swin\cite{Liu2021SwinTH} and CSWin\cite{Dong2021CSWinTA} as backbone for feature extraction, (2) Train on full GLDv2~\cite{Weyand2020GoogleLD}, and (3) Using full GLDv2~\cite{Weyand2020GoogleLD} images as index image set for kNN search.

With these modifications, our solution significantly improves last year solution on this year competition. Our full pipeline, after ensembling Swin\cite{Liu2021SwinTH}, CSWin\cite{Dong2021CSWinTA}, EfficientNet\cite{Tan2019EfficientNetRM} B7 models, scores 0.4907 on the private leaderboard which help us to get the 2nd place in the competition. 
\end{abstract}

\maketitle

\section{Introduction}
Google Landmark Recognition 2021 Competition~\cite{landmark-recognition-2021} is the fourth landmark Recognition competition on Kaggle. The task of image Recognition is to build models that recognize the correct landmark (if any) in a dataset of challenging test images. This year, hosts introduced a lot more diversity in the challenge’s test images in order to measure global landmark recognition performance in a fairer manner, and following last year’s success,the competition is set as a code competition which emphasis building more efficient model and generalizing to unseen test set. Google Landmarks Dataset v2(GLDv2)~\cite{Weyand2020GoogleLD} is the biggest landmark dataset, which contains approximately 5 million images, split into 3 sets of images: train, index and test. There are 4132914 images in train set, 761757 images in index set. To facilitate recognition-by-retrieval approaches, the private training set contains only a 100k subset of the total public training set.This 100k subset contains all of the training set images associated with the landmarks in the private test set. Competitors may still attach the full training set as an external data set. Submissions are evaluated using Global Average Precision (GAP) at k, where k=1. 

\section{Methods}


\begin{table*}[h]
  \begin{tabular}{lcccccc}
    \toprule
     Models/Methods & Image Size & Post-Processing & train on GLDv2 & Index on GLDv2 & public & private\\
    \midrule
    EfficientNet B7	                         &800&-      &-		 &-		&0.375&0.359\\
    EfficientNet B7	                         &736&$\surd$&-		 &- 	&0.424&0.414\\
    SwinBase	                             &736&$\surd$&-      &-     &0.451&0.427\\
    EfficientNet B7	                         &736&$\surd$&$\surd$&-     &0.419&0.397\\
    SwinBase	                             &736&$\surd$&$\surd$&-      &0.463&0.434\\
    EfficientNet B7                          &736&$\surd$&$\surd$&$\surd$&0.451&0.439\\
    SwinBase	                             &736&$\surd$&$\surd$&$\surd$&0.494&0.464\\
    SwinLarge	                             &800&$\surd$&$\surd$&$\surd$&0.454&0.448\\
    CSWinLarge	                             &736&$\surd$&$\surd$&$\surd$&0.414&0.429\\
    SwinBase+SwinLarge		     &-  &$\surd$&$\surd$&$\surd$&0.508&0.482\\
    SwinBase+SwinLarge+CSWinLarge+B7	 &-  &$\surd$&$\surd$&$\surd$&0.517&0.490\\
    \bottomrule
  \end{tabular}
  \caption{ Leaderboard Scores of Google Landmark Recognition 2021. }
  \label{tab:bestfitting_result}
\end{table*}

\subsection{Retrieval Models}
To extract global descriptors of landmark images, Transformer-based architectures were tested and employed as backbone, they showed much better results than CNN based architectures such as EfficientNet\cite{Tan2019EfficientNetRM} B7, Resnet152~\cite{He2016DeepRL} used in  last year's solution.

\subsubsection{\textbf{Model Design}}

Three types of models were used for final submission and each model includes a backbone for feature extraction and head layers for classification. These backbones are SWIN\cite{Liu2021SwinTH}, CSWIN\cite{Dong2021CSWinTA},  EfficientNet\cite{Tan2019EfficientNetRM} B7 was also kept to add diversity to our solution. 

Head layers includes a pooling layer and two fully connected(fc) layers. We use generalized mean-pooling (GeM)~\cite{Gu2018AttentionawareGM} as pooling method since it has superior performance and p of GeM is set to 3.0 and fixed during the training. The first fc layer is often called embedding layer whose output size is 512, and we will extract the output of this layer as global description of an image. While the output size of the second fc layer is corresponding to the class number of training dataset(81k and 200k in different stages). Instead of using softmax loss for training, we train these models with arcmargin loss~\cite{Deng2019ArcFaceAA}, the arcmargin-scale is set to 30,  arcmargin-margin is set to 0.3. For validation set, sample 200 images from GLDv2 test set as val set and all the ground truth images of Google Landmark Retrieval Competition 2019 as index dataset and calculated the mAP@100 score. Despite the small sized validation set, the score correlated well with the leaderboard score.

\subsubsection{\textbf{Training Details}}
We trained our models by increasing image size step by step following the strategy of 1st place solution~\cite{Jeon20201stPS} to Google Landmark Retrieval 2020 with some modifications. 

\begin{itemize}
\item[$\bullet$] \textbf{Stage 1}, cleaned GLDv2 was used to train the model to classify 81313 landmark classes. SWIN\cite{Liu2021SwinTH} and CSWIN\cite{Dong2021CSWinTA} backbone based model was trained 6 epochs with 448×448 image inputs at this step.

\item[$\bullet$] \textbf{Stage 2}, as there are 3.2 million images belong to the 81313 classes in cleaned GLDv2. we defined these 3.2m images as GLDv2x. GLDv2x was used to finetune the model from step 1 for 4 epochs.

\item[$\bullet$] \textbf{Stage 3}, model from step 2 was finetuned using 512×512 images from GLDv2x for 6 epochs.

\item[$\bullet$] \textbf{Stage 4}, finetuned with 640×640 images for 3 epochs on GLDv2x.

\item[$\bullet$] \textbf{Stage 5}, finetuned with 736×736 images for 3 epochs on GLDv2x.

\item[$\bullet$] \textbf{Stage 6}, finetuned with 736×736 images for 3 epochs on full GLDv2 dataset which have more than 200k classes.

\item[$\bullet$] \textbf{Stage 7}, finetuned with 800×800 images for 1 epoch on full GLDv2 dataset.

\end{itemize}

Stochastic gradient descent optimizer was used for training, where learning rate, momentum, weight decay are set to 1e-2, 0.9, 1e-5. learning rate was set to 0.001 for last 2 epochs. For image augmentation, left-right flip was used when image size is 448×448. When our models were finetuned on larger images, we used some complex augmentations, including RandomCrop, Brightness, Color, Cutout, Contrast, Shear, Translate, Rotate90.

\subsection{Validation Strategy} 
\begin{itemize}
\item[$\bullet$] \textbf{The val set part 1}: the 1.3k landmark images from GLDv2 test set( exclude those not in 81k classes).\\
\item[$\bullet$] \textbf{The val set part 2}: sample 2.7k images from GLDv 2x but not in cleaned GLDv2.\\
\item[$\bullet$] \textbf{The index image set for val set}: all the images of GLDv2x.\par
\end{itemize}

The difference of this strategy from last year's is the index image set is change from part of the GLDv2x to full GLDv2x.

\subsection{Data Usage Strategies}

We found finetuning on GLDv2 with more than 200k classes can help to improve the score on public LB. As shown in Table 1, after finetuned on GLDv2 full dataset, Swin Base model improved from 0.451/0.427 to 0.463/0.434 on public and private LB.

Further experiment results showed that uploading full GLDv2 features as index image set features for kNN search can improve the score significantly, as listed in Table 1, the Swin Base model improved from 0.463/0.434 to 0.494/0.464.

\subsection{Spatial verification and Post-Processing}

As opposed to last year's solution, spatial verification was not used in this competition by two reason: the first, the score did not improved on public leaderboard, and the second, it's infeasible do it on full GLDv2.

The post-processing strategy is almost the same as last year's solution, but we did not use spatial verification results as features.

The competition metric is  Global Average Precision (GAP), if non-landmark images (distractors) are predicted with higher confidence score than landmark images. Hence, it is essential to suppress the prediction confidence score of these distractors.

We tried rules from winner solutions of Google Landmark Recognition Competition 2019~\cite{landmark-recognition-2019}~\cite{Ozaki2019LargescaleLR}~\cite{Chen20192ndPA}~\cite{Gu2019TeamJS}, the following are the effective rules:\par
1.Search top 3 non-landmark images from no-landmark image set for query an image, if the similarity of top3\textgreater0.3, then decrease the score of the query image~\cite{Chen20192ndPA}.\par
2.If a landmark is predicted\textgreater20 times in the test set, then treat all the images of that landmark as non-landmarks~\cite{Ozaki2019LargescaleLR}.\par

As many features(similarity to index images, similarity to non-landmark images etc.) can be used for determining whether an image is non-landmark or not, we developed a model which can be called re-rank model following the re-rank strategy in tweet sentiment extraction ~\cite{re-ranking-candidates}, the difference is that we use tree model instead of NN model. 

\section{Ensemble}
We used 4 models in our final ensemble, they are SwinBase, SwinLarge, CSWinLarge and EfficientNet B7, they all finetuned on GLDv2 full dataset. We extracted features from images in full GLDv2 and uploaded to serve as index dataset for recognition-by-retrieval approach. To add diversity, we extracted features using different image sizes and several Test Time Augmentations. There are 11 types of features are used in our final ensemble, which help us achieve 0.517/0.490 on public and private leaderborad.

\section{Conclusion}
In this paper, We presented a detailed solution for the Google Landmark Recognition 2021. The solution improved from last year's solution by employing transformer-based models as backbone and using metric learning models which trained step by step on bigger and larger images to find candidate images from index set for a query image, then get landmark id and score by probability of classification and similarity between the query image and index images. To suppress distractors, the post-processing rules and models played a important role again in our final pipeline.



\bibliographystyle{ACM-Reference-Format}
\bibliography{sample}

\end{document}